\pdfoutput=1
\documentclass[letterletter, 10 pt, conference]{ieeeconf}  

\IEEEoverridecommandlockouts                              

\title{\LARGE \bf
Safe\text{-}VLN: Collision Avoidance for Vision-and-Language Navigation of Autonomous Robots Operating in Continuous Environments
}

\author{Lu Yue, Dongliang Zhou, Liang Xie, Feitian Zhang, Ye Yan, and Erwei Yin 
\thanks{Lu Yue is with the College of Engineering, Peking University, Beijing, 100871, China, Defense Innovation Institute, Academy of Military Sciences, Beijing 100071, China, and Tianjin Artificial Intelligence Innovation Center, Tianjin 300450, China (email: yuelu@stu.pku.edu.cn).}%
\thanks{Dongliang Zhou is with Harbin Institute of Technology, Shenzhen, Shenzhen University Town (HIT Campus), Shenzhen 518055, China.}%
\thanks{Feitian Zhang is with the Department of Advanced Manufacturing and Robotics, and the State Key Laboratory of Turbulence and Complex Systems, College of Engineering, Peking University, Beijing, 100871, China.
        }%
\thanks{Liang Xie, Ye Yan, and Erwei Yin are with Defense Innovation Institute, Academy of Military Sciences, Beijing 100071, China, and also with Tianjin Artificial Intelligence Innovation Center, Tianjin 300450, China. 
        }
\thanks{Corresponding authors: Feitian Zhang, email: feitian@pku.edu.cn, and Erwei Yin, email: yinerwei1985@gmail.com.}    
        }
\usepackage{soul} 
\usepackage{graphicx} 
\usepackage{amsmath}
\usepackage{multirow}
\usepackage{makecell}
\usepackage{bm}
\usepackage{lineno}
\usepackage{pifont}
\usepackage{bigstrut}
\usepackage[table]{xcolor}

\begin{document}


\maketitle
\pagestyle{empty}  
\thispagestyle{empty} 
\thispagestyle{empty}
\pagestyle{empty}

\begin{abstract}
The task of vision-and-language navigation in continuous environments (VLN-CE) aims at training an autonomous agent to perform low-level actions to navigate through 3D continuous surroundings using visual observations and language instructions.
The significant potential of VLN-CE for mobile robots has been demonstrated across a large number of studies.
However, most existing works in VLN-CE focus primarily on transferring the standard discrete vision-and-language navigation (VLN) methods to continuous environments, overlooking the problem of collisions. Such oversight often results in the agent deviating from the planned path or, in severe instances, the agent being trapped in obstacle areas and failing the navigational task. To address the above-mentioned issues, this letter investigates various collision scenarios within VLN-CE and proposes a classification method to predicate the underlying causes of collisions. 
Furthermore, a new VLN-CE algorithm, named Safe-VLN, is proposed to bolster collision avoidance capabilities including two key components, i.e., a waypoint predictor and a navigator. In particular, the waypoint predictor leverages a simulated 2D LiDAR occupancy mask to prevent the predicted waypoints from being situated in obstacle-ridden areas. The navigator, on the other hand, employs the strategy of `re-selection after collision' to prevent the robot agent from becoming ensnared in a cycle of perpetual collisions. The proposed Safe-VLN is evaluated on the R2R-CE, the results of which demonstrate an enhanced navigational performance and a statistically significant reduction in collision incidences.

\end{abstract}


\begin{figure}
\centering 
\includegraphics[width=1.0\linewidth]{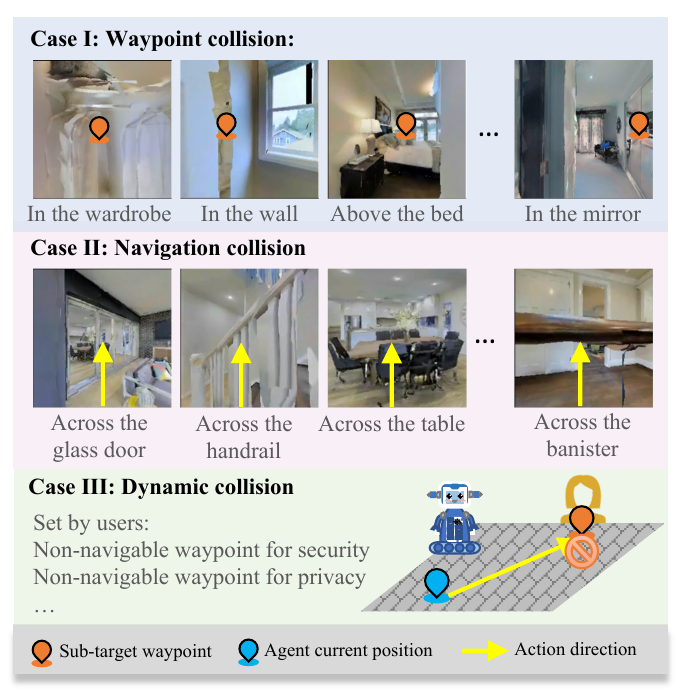}
\vspace{-0.6cm}
\caption{Collision scenarios in VLN-CE classified in this letter. The first type of collision refers to the predicted waypoint being in the obstacle zone; the second type of collision refers to encountering obstacles when navigating from the waypoint to the next chosen waypoint; and the third type of collision refers to dynamic obstacles that may occur at any time during the navigation process.}
\label{collision}
\vspace{-0.7cm}
\end{figure}
\section{Introduction}\label{introduction}

In recent years, autonomous navigation \cite{hu2021sim,kahn2021badgr} of mobile robots has been experiencing a surge of interest due to its immense commercial potential and multifaceted application value. A specific task within this realm, referred to as vision-and-language navigation (VLN) \cite{anderson2018vision} has garnered considerable scholarly attention. 
VLN aims to train an agent to navigate with cross-modal matching information of visual data and natural language instructions by interacting with discrete environments through connectivity graphs. Nevertheless, most of the previous studies \cite{hong2020recurrent,  cui2023grounded} are concerned about discrete VLN configurations that simplify navigation to the task of traversing environments with pre-defined graphs. 
In particular, these pre-defined graph settings inherently introduce prior assumptions concerning navigational locations and perfect navigation between nodes.
This leads to a significant challenge when attempting to apply VLN models trained in simulations to real-world systems.

To bridge the gap between VLN and the real-world navigation, Krantz \textit{et al.} \cite{krantz2020beyond} first introduced VLN in continuous environments (VLN-CE).
They instantiated VLN within a 3D continuous environment simulator, known as Habitat \cite{savva2019habitat}.
In particular, VLN-CE eliminates the necessity for discrete graph assumptions and opts for a more generalized setting of navigation within a continuous environment, bringing the application of VLN significantly closer to real-world scenarios.
In comparison to its discrete counterpart, VLN-CE faces greater challenges in its design. For instance, the incorporation of low-level actions considerably escalates the algorithmic complexity, thereby slowing the convergence of the learning process, particularly in end-to-end training.

To learn effectively from long-horizon trajectories, the mainstream framework for VLN-CE comprises three modules \cite{krantz2022sim}, including a waypoint predictor, a navigation planner, and a low-level controller. 
Specifically, learning from the known navigation graphs in Matterport3D, the waypoint predictor \cite{krantz2022sim, anderson2021sim} generates high-level navigation waypoints in continuous environments by predicting nearby candidate positions from visual observations. 
Based on the pre-trained waypoint predictor, the navigation planner generates the next subgoal waypoints that are further implemented by the low-level controller. 
While certain studies focus on improving the performance of VLN-CE  \cite{krantz2021waypoint, irshad2022semantically}, very few has considered the collision avoidance and navigation safety. As shown in Fig. \ref{collision}, collisions occur frequently in VLN-CE, mainly facing the following challenges. First,
in comparison to those in discrete environments, the waypoints generated by the waypoint predictor in continuous environments may not be navigable due to the incorrect perception of surrounding environments. Second,
without the perfect navigation between adjacent nodes, the agent more likely fails to reach the next waypoint in continuous environments when involving obstacles along the travelling path.
To enhance collision avoidance of VLN-CE, this letter takes a thorough and quantitative analysis of VLN-CE collisions and proposes a new navigation algorithm, named Safe-VLN. Specifically, this letter investigates causes of collisions in VLN-CE and defines three different types of collisions. Designed to resolve all the types of collisions, the Safe-VLN is expected to greatly improve collision avoidance capability of the majority of VLN-CE agents. 

The contributions of this letter are twofold.
First, to the best of the authors' knowledge, this letter classifies, for the first time, the VLN-CE collisions into three types, including the waypoint collision, the navigation collision and the dynamic collision. Through extensive experimentation, this letter quantitatively investigates performance degradation caused by different types of collisions, providing valuable insights for the design of collision avoidance algorithms.
Second, this letter proposes a new Safe-VLN algorithm to address the collision issues in VLN-CE. Experimental results validate the effectiveness of the proposed algorithm, which successfully achieves a state-of-the-art success rate in VLN-CE on the widely-used R2R-CE dataset.



The remainder of this letter is organized as follows. Section \ref{related work} reviews related work. Safe-VLN is proposed in Section \ref{Safe-VLN} and Section \ref{experiment} details dataset construction and presents the experimental results. Finally, Section \ref{conclusion} provides the concluding remarks.

\section{Related Work}\label{related work}
In this section, we review related research on VLN-CE and collision avoidance in autonomous navigation. We highlight the features of this letter in comparison to previous studies.

\textbf{VLN-CE.}
To be more representative of real-world navigation, Krantz  \textit{et al}. \cite{krantz2020beyond} converted the discrete VLN task into a continuous-environment task.
To avoid slow convergence brought by end-to-end baselines \cite{krantz2021waypoint}, Krantz \emph{et al.} \cite{krantz2022sim}, Hong \emph{et al.} \cite{hong2022bridging}, Wang \emph{et al.} \cite{gridmm} and An \emph{et al.} \cite{an2023etpnav} adopted a hierarchical framework by training a waypoint predictor module to obtain local candidate waypoints. This hierarchical framework is commonly used in VLN-CE and further extended with top-down
maps for better perception of environments, i.e., the top-down egocentric and dynamically growing grid memory map is built by GridMM to represent the previously visited environment \cite{gridmm}. However, navigation methods 
present a significantly lower navigational performance in VLN-CE than in VLN. 

An \emph{et al.} \cite{an2023etpnav} reported that in continuous environments, without the perfect navigation assumption, collisions often occur, causing the agent to deviate from the planned route or even fall into deadlocks. Based on this, this letter analyzes and defines collisions in VLN-CE. To transfer a VLN agent to continuous environments with collision considerations, Safe-VLN is proposed to improve the VLN-CE by adding the occupancy mask in waypoint predictor module and employing `re-selection' strategy in navigation planner.

\textbf{Collision in Autonomous Navigation.} 
There exist many collision avoidance studies in robot navigation. One of the most commonly-used methods is the traditional grid map-based path planning algorithm
such as the dual Dijkstra search algorithm \cite{collision1}, the A* algorithm \cite{collision2}, and the jump point search (JPS) algorithm \cite{collision3}. 
However, given the VLN-CE task setting, applying these collision avoidance methods requires the extraction of grid maps from the original image observations that are directly modeled from real-world indoor environments.
This operation not only increases the computational complexity but also scarifies environmental information and introduces more localization errors, thus potentially degrading the obstacle avoidance performance.

Over the past several years, deep reinforcement learning (DRL) has been widely researched and applied as a contemporary design method for collision avoidance \cite{learning1, learning2, learning3}.
While DRL has been proven as an effective collision avoidance method in numerous navigation tasks, the application in VLN-CE is still an open
question with great challenges. The main reason is that compared with other
widely-studied navigation tasks, VLN-CE is extremely complex and difficult considering the multi-modal
inputs with simulation environments directly modeled from the real world. The integration
of DRL with VLN-CE will significantly increase the computational time due to the extensive
exploration in a vast search space, thus leading to an 
extremely slow
learning convergence.

ETPNav, a recent work that considers collisions in VLN-CE \cite{an2023etpnav} leverages a heuristic strategy called `Tryout' in the low-level controller to avoid the agent from being stuck in collision areas by randomly selecting actions from a predefined set of actions. Nevertheless, in several cases, the agent still gets trapped after trying all the actions in the set. Tuning the low-level control is not enough to solve the complex collision problems in VLN-CE. 
Inspired by the collision studies in autonomous navigation, we categorize the causes of collisions in VLN-CE and propose Safe-VLN with improvements in the waypoint predictor and the navigation planner accordingly.


\begin{figure}[t]
\centering 
\includegraphics[width=1\linewidth]{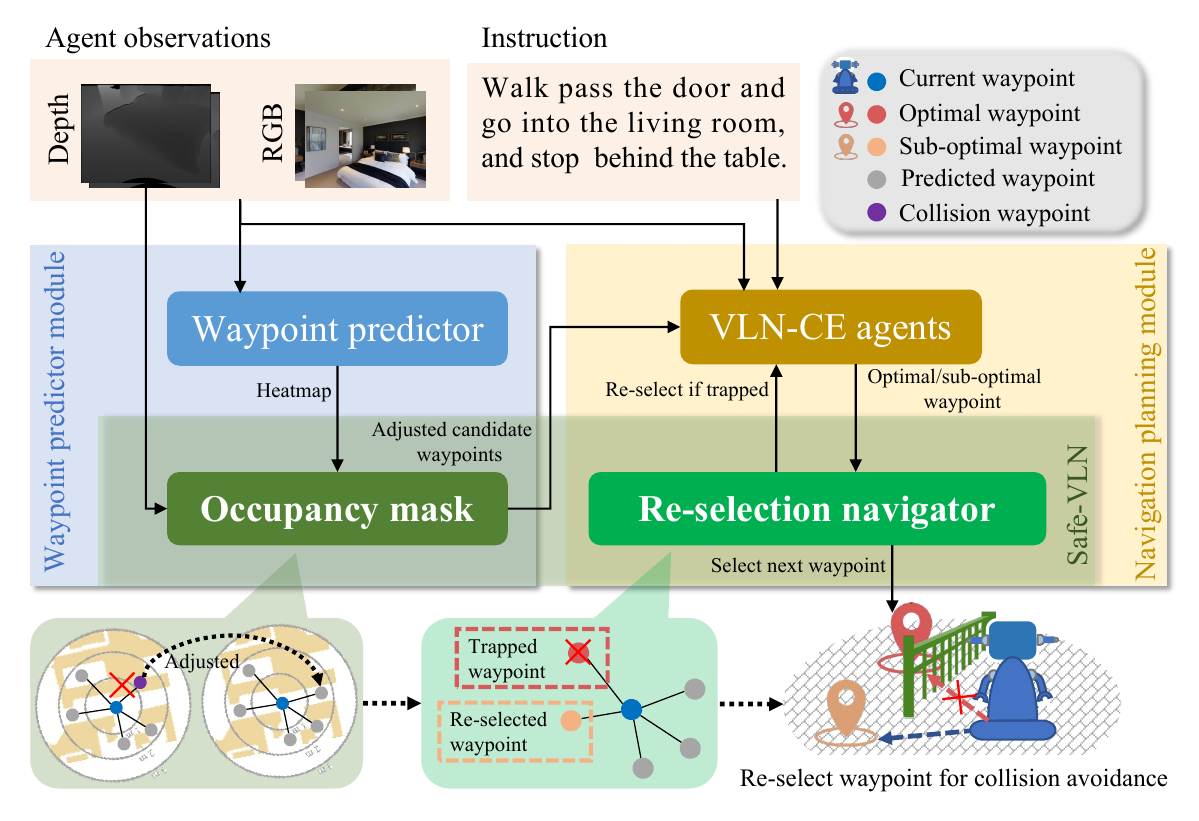}
\vspace{-0.7cm}
\caption{The overview of the proposed Safe-VLN. The Safe-VLN designs a waypoint predictor that generates candidate waypoints by combining the predicted heatmap with occupancy masks. In addition, the Safe-VLN adds a re-selection navigator in the navigation planner.}
\label{main_idea}
\vspace{-0.6cm}
\end{figure}

\section{Safe-VLN}
\label{Safe-VLN}
In this section, we first present the general navigation setups in VLN-CE and then conduct an analysis of various collision scenarios and subsequently categorize them into three distinct types. Following this, we propose Safe-VLN, a strategy designed to mitigate these collision types for agents operating in continuous environments. This approach incorporates occupancy masks into the waypoint predictor and employs a `re-selection' planning strategy.

\subsection{Overview of VLN-CE}
For the sake of self-containment, this letter includes a general overview of VLN-CE.
In line with previous studies \cite{krantz2022sim, hong2022bridging, an2023etpnav}, the agent in VLN-CE is provided with an instruction composed of $L$ words for each episode. This instruction is denoted as $I = \{l_1,\cdots,l_L\}$ with associated text embeddings. 
During the navigation phase, at each time step $t$, the agent receives visual observations $O_t=\{(o^{\text{rgb}}_{t,i}, o^{\text{d}}_{t,i})\}_{i=1}^{12}$, where $o^{\text{rgb}}_{t,i}, o^{\text{d}}_{t,i}$ are the $i$-th RGB and depth images, respectively. These images are captured by the mobile agent across 12 different views, each with an equally-spaced horizontal heading angle relative to the agent's orientation, represented by $\theta_i = 30^{\circ}\times(i-1)$, where $i=1,\cdots,12$.
Based on the visual observations $O_t=\{(o^{\text{rgb}}_{t,i}, o^{\text{d}}_{t,i})\}_{i=1}^{12}$, two separate pre-trained image encoders are utilized to extract RGB features $V^{\text{rgb}}_t=\{v^{{\text{rgb}}}_i\}^{12}_{i=1}$ and depth features $V^{\text{d}}_t=\{v^{\text{d}}_i\}^{12}_{i=1}$. In particular, 
a vision transformer (ViT) \cite{radford2021learning} derived from contrastive language-image pre-training (CLIP) \cite{dosovitskiy2020image} is employed to encode the RGB images, while a residual network \cite{he2016deep, zhou2022coutfitgan} pre-trained for point-goal navigation \cite{wijmans2019dd} is used to encode the depth images.
Leveraging instruction embeddings and visual observation features, the agent is expected to approach and ultimately reach the target as dictated by the instruction. Unlike the selection of discrete waypoint candidates, agents in VLN-CE navigate through a 3D meshed environment with continuous values by taking actions from a lower-level action space, e.g., {forward 0.25m, turn-left 15°, turn-right 15°, and stop}. 
The navigation configurations in VLN-CE abandon the graph prior and the assumption of perfect navigation between nodes, making the task more challenging. To address the collision challenges in VLN-CE, we introduce Safe-VLN to enhance the navigation performance of the agent when adapting to complex continuous environments.

\subsection{Collision Classification in VLN-CE}
This subsection presents an analysis of various collision scenarios, and defines three types of collisions pertinent to tasks within VLN-CE. As depicted in Fig. \ref{collision}, 
the first type is named `waypoint collision'. This category of collisions occurs when the predicted candidate waypoint is positioned within an obstacle area, such as inside a wardrobe, on the wall, or above a bed, making it impossible to reach the exact waypoint without a collision.
The second type, `navigation collision', is defined as a collision that occurs when the agent encounters obstacles like handrails, doors, tables, etc., during its movement towards the next waypoint. The third type, `dynamic collision', occurs when a waypiont becomes non-navigable during the process of robot motion due to time-varying security or privacy concerns instead of static obstacles.
Upon closer examination of VLN-CE algorithms, such as CWP-RecBERT \cite{hong2022bridging}, GridMM \cite{gridmm} and ETPNav \cite{an2023etpnav}, it becomes evident that none of these methods address the three types of collisions defined in this research.
Notably, while `dynamic collision' is not presented in the existing dataset for VLN-CE, it is nonetheless considered in this research, in order to comprehensively study and resolve all the types of collisions in real-world applications. to simulate the situation of `dynamic collision' in real-world scenarios, we generate the next waypoint selected by the agent as non-navigable with a certain probability.
In particular, validation results in Table \ref{table1} of Section IV from the `val-unseen' split of the R2R-CE dataset reveal that
both waypoint collisions and navigation collisions occur frequently in existing algorithms, and their success rates decrease significantly when dynamic collisions are considered.


\subsection{Occupancy Masks for Waypoint Collision}
The first module of Safe-VLN is designed to alleviate the waypoint collisions. With a simulated 360° 2D LiDAR scanner, an occupancy mask is created with surrounding environment information and further incorporated into the waypoint predictor of VLN-CE.
Specifically, we use the waypoint predictor in ETPNav \cite{an2023etpnav} to generate a probability heatmap $H_t$, denoting probabilities of nearby candidate waypoints within a 3m radius centered around the agent. The heatmap $H_t$ is modeled as
\vspace{-1mm}
\begin{equation}
\label{heatequa}
		H_t = \text{WP}(V^{\text{d}}_t, V^{\text{ori}}_t),
\end{equation}
\vspace{-1mm}
where WP$(\cdot)$ represents the transformer-based waypoint predictor, and $V^{\text{d}}_t=\{v^{\text{d}}_i\}^{12}_{i=1}$, $V^{\text{ori}}_t=\{(\cos{\theta_i}, \sin{\theta_i})\}^{12}_{i=1}$ are depth and orientation features, respectively. 
$K$ predicted candidate waypoints are sampled from the heatmap $H_t$ using the non-maximum-suppression (NMS) technique and the waypoint positions are denoted by a set $P_t = \{p^k_t\}^K_{k=1}$.

The aforementioned waypoint predictor, trained based on the navigation graphs in the Matterport3D dataset, does not account for the surrounding obstacles. As a result, the predicted waypoints occasionally locate within obstacle areas, leading to waypoint collision failures. To address this issue, this research employs depth information to generate a simulated 2D-LiDAR occupancy mask $M_t$ defined by
\vspace{-0.5mm}
\begin{equation}
M_t(h,w) =\begin{cases}
        -1 ,& \text{if} \; M_t(h,w) \; \text{is occupied by obstacles}; \\
        0 ,& \text{otherwise}.
     \end{cases}
\end{equation}
\vspace{-0.5mm}
Here, $M_t$ shares the same dimensions $(h,w)$ as the predicted heatmap $H_t$.

\begin{figure}[t]
\centering 
\includegraphics[width=0.9\linewidth]{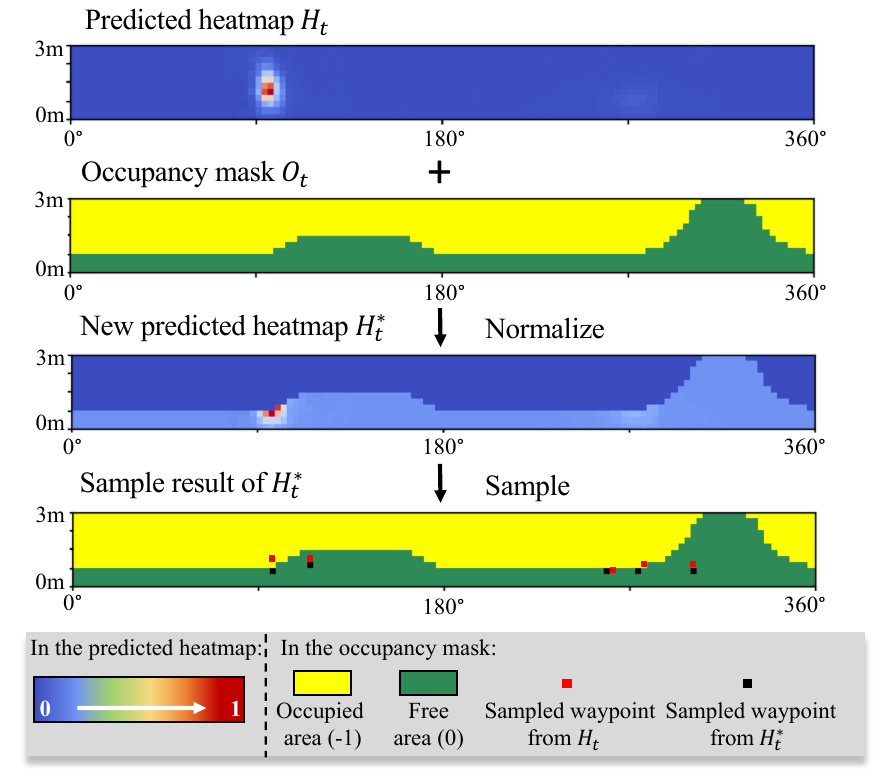}
\vspace{-0.2cm}
\caption{Illustration of occupancy masks designed in the waypoint predictor. }
\label{heat-map}
\vspace{-0.6cm}
\end{figure}

As shown in Fig. \ref{heat-map}, the occupancy mask $M_t$, centered on the agent, provides ranging information of surrounding environments.
By incorporating the occupancy mask into the predicted heatmap $H_t$, the probability within collision areas in $H_t$ is reduced, thereby obtaining an updated heatmap $H^*_t$
\vspace{-1mm}
\begin{equation}
\label{eq5}
        H^*_t = \text{norm}(H_t + \delta \cdot M_t),
\end{equation}
\vspace{-1mm}
where $\text{norm}(\cdot)$ is a normalization operation, ensuring that the sum of probability values over the updated heatmap equals one, and $\delta$ is a hyper-parameter to control the weights of the mask in the waypoint prediction.

\subsection{Re-selection Navigator for Navigation and Dynamic Collisions}
In addition to the waypoint collisions, the agent frequently encounters navigation and dynamic collisions. To address these two types of collisions, a re-selection navigator is proposed as the second module of Safe-VLN. This solution involves designing the navigation planner to predict both the optimal and sub-optimal waypoints for alternatives. When the agent is trapped, the re-selection of sub-optimal waypoints not 
only help it escape from the collision area but also prevents significant deviation from the target.

Analogous to ETPNav, to achieve long-term planning, the traversed path for each episode is depicted by a graph, symbolized as $G_t=(N_t, E_t)$. Here, $N_t = \{n_t^i\}_{i=0}^{|N_t|}$ denotes the set of nodes, which includes visited nodes, the current node, and observed but not visited `ghost nodes'. The special node, $n_0$, is interconnected with all other nodes, symbolizing the action `stop'. $E_t$ represents the set of all the edge elements $e_t^{i,j}$, where $e_t^{i,j}$ encapsulates the relative Euclidean distance between two adjacent nodes $n_t^i$ and $n_t^j$. $G_t$ is updated at each $t$ by the predicted waypoints.
For instance, the embedding for each node, $n_t^i$, is represented by $\hat{n}^i_t=V^{\text{rgb}}_t\oplus V^{\text{d}}_t\oplus V^{\text{pose}}_t\oplus  V^{\text{step}}_t$, where $\oplus$ denotes a concatenation operation. Here, $V^{\text{pose}}_t$ is the current pose embedding of the agent and $V^{\text{step}}_t$ denotes the embedding of the most recently visited time step. The visual-semantic embedding, $\widetilde{n}_t^i$, for each node, $n_t^i$, is obtained by
\vspace{-1mm}
\begin{equation}
	\widetilde{n}_t^i = \text{Cross-Modal}(I, \hat{n}_t^i),
\end{equation}
\vspace{-1mm}
where Cross-Modal$(\cdot)$ is a cross-modal planning model containing a multi-layer transformer to compute the navigation plan based on graph features and navigation instructions.
The navigation planning policy generates scores, $s_t=\{s_t^i\}_{i=0}^{|N_t|}$, for nodes in $G_t$ using the following form, which reflects the probability preference for selecting each node
\begin{equation}
		s_t^i = \text{NP}(\widetilde{n}_t^i), i=0,...,|N_t|,
\end{equation}
where NP$(\cdot)$ is the navigation planning policy.
Additionally, the scores for visited nodes and the current node are masked to prevent unnecessary repetition of actions. Based on the output scores, $s_t$, the agent selects the next waypoint in a greedy manner. If $s_t^0$ is selected, the agent will cease movement at its current position, resulting in the termination of the current episode.

The training objective of the existing VLN-CE models such as ETPNav is to maximize the scores of the optimal waypoint, which is characterized by the shortest Dijkstra path to the target. To generate appropriate alternative waypoints, the sub-optimal waypoint, characterized by the second shortest path, is also considered in the training objective by updating the loss function as follows
\begin{equation}
\label{equ6}
		\mathcal{L}_p = \sum_{t}[\lambda_1\log p(a^1_t|\mathbf{W},G_t)+\lambda_2\log p(a^2_t|\mathbf{W},G_t)],
\end{equation}
\vspace{-1mm}
where $a^1_t$ and $a^2_t$ are the optimal and sub-optimal actions with the shortest and second shortest paths to the target, respectively. Based on the graph $G_t$ and the network parameters $\mathbf{W}$ at time step $t$, the selection probabilities of $a^1_t$ and $a^2_t$ are represented as $p(a^1_t|\mathbf{W},G_t)$ and $p(a^2_t|\mathbf{W},G_t)$. $\lambda_1$ and $\lambda_2$ are 
hyper-parameters for tradeoff between optimal and sub-optimal actions.

\begin{table}[hb]
\vspace{-0.2cm}
  \centering
  \caption{Validation performance of different VLN-CE agents on the $\emph{val-unseen}$ dataset with or without Safe-VLN. Here, for all metrics except W-C and N-C, higher is better}
  \vspace{-0.2cm}
  \setlength\tabcolsep{1.2mm}{
    \begin{tabular}{c|c|c|c|c|c|c}
    \Xhline{1.25pt}
    Method & Safe-VLN & SR    & SPL   & W-C   & N-C   & D-C SR \bigstrut\\
    \hline
    \multirow{2}[2]{*}{CWP-RecBERT \cite{hong2022bridging}} & \ding{55}      & 44.0  & 38.9  & 11.3  & 11.5  & 40.5  \bigstrut[t]\\
          &  \ding{51}\cellcolor{gray!40}     & 44.9\cellcolor{gray!40}  & 39.2\cellcolor{gray!40}  & 10.2\cellcolor{gray!40}  & 5.8\cellcolor{gray!40}   & 41.9\cellcolor{gray!40}  \bigstrut[t]\\
    \hline
    \multirow{2}[2]{*}{GridMM \cite{gridmm}} & \ding{55}      & 48.5  & 40.7  & 11.3  & 8.2   & 46.3 \bigstrut[t]\\
          &  \ding{51}\cellcolor{gray!40}     & 49.4\cellcolor{gray!40}  & 41.2\cellcolor{gray!40}  &10.2\cellcolor{gray!40}  &4.4\cellcolor{gray!40}   &47.3\cellcolor{gray!40}  \bigstrut[t]\\
    \hline
    \multirow{2}[2]{*}{ETPNav \cite{an2023etpnav}} & \ding{55}      & 57.1  & 49.0  & 23.3  & 14.0  & 54.1  \bigstrut[t]\\
          &  \ding{51}\cellcolor{gray!40}     & 57.9\cellcolor{gray!40}  & 47.8\cellcolor{gray!40}  & 22.8\cellcolor{gray!40}  & 4.3\cellcolor{gray!40}   & 54.3\cellcolor{gray!40}  \bigstrut[t]\\
    \Xhline{1.25pt}
    \end{tabular}%
    }
    \label{table1}%
  \vspace{-0.6cm}
\end{table}%

\begin{table*}[htbp]
\vspace{-0.3cm}
  \centering
  \caption{Comparison with state-of-the-art methods on the R2R-CE dataset}
    \vspace{-0.2cm}\begin{tabular}{c|c|c|c|c|c|c|c|c|c|c|c|c|c|c|c}
    \Xhline{1.25pt}
    \multirow{2}[4]{*}{Method} & \multicolumn{5}{c|}{Val-Seen}                 & \multicolumn{5}{c|}{Val-Unseen}               & \multicolumn{5}{c}{Test Unseen} \bigstrut\\
\cline{2-16}          & TL   & NE    & OSR   & SR  & SPL    & TL   & NE     & OSR    & SR  & SPL     & TL    & NE    & OSR    & SR  & SPL  \bigstrut\\
    \hline
    HPN+DN \cite{krantz2021waypoint} & 8.54  & 5.48  & 53    & 46    & 43      & 7.62  & 6.31  & 40    & 36    & 34       & 8.02  & 6.65  & 37    & 32    & 30 \bigstrut\\
    SASRA \cite{irshad2022semantically} & 8.89  & 7.71  & -     & 36    & 34       & 7.89  & 8.32  & -     & 24    & 22       & -     & -     & -     & -     & - \\
    Sim2Sim \cite{krantz2022sim} & 11.18 & 4.67  & 61    & 52    & 44        & 10.69 & 6.07  & 52    & 43    & 36        & 11.43 & 6.17  & 52    & 44    & 37 \\
    CWP-CMA \cite{hong2022bridging} & 11.47 & 5.20   & 61    & 51    & 45      & 10.90  & 6.20   & 52    & 41    & 36     & 11.85 & 6.30   & 49    & 38    & 33 \\
    CWP-RecBERT \cite{hong2022bridging} & 12.50  & 5.02  & 59    & 50    & 44    & 12.23 & 5.74  & 53    & 44    & 39    & 13.51 & 5.89  & 51    & 42    & 36 \\
    GridMM \cite{gridmm} & 12.69  & 4.21  & 69    & 59    & 51    & 13.36 & 5.11  & 61    & 49    & 41    & 13.31 & 5.64  & 56    & 46    & 39 \\
    ETPNav \cite{an2023etpnav} & 11.92 & 3.96  & 72    & 66    & 58     & 11.97 & 4.71  & 64    & 57    & 49     & 12.87 & 5.12  & 63    & 55    & 48 \bigstrut\\ \rowcolor{gray!40}
    \hline
Safe-VLN (ours) & 13.71 & 3.35      & 79    & 71    & 60      & 15.00 & 4.48  & 68    & 60    & 47      &  15.44     & 5.01      & 64      & 56      & 45 \bigstrut\\
    \Xhline{1.25pt}
    \end{tabular}%
  \label{table3}%
  \vspace{-0.5cm}
\end{table*}%

\section{Experiments}\label{experiment}

In this section, we initially describe the experimental setup, encompassing the dataset and training setup. Subsequently, we compare our approach with a classical obstacle avoidance algorithm and conduct experiment by comparing the Safe-VLN with other baselines in VLN-CE. Following this comparison, ablation experiments are carried out to verify the impact of sensor errors and validate the effectiveness of each module within Safe-VLN.

\subsection{Experimental Setup}

\textbf{Dataset.}
Matterport3D (MP3D) \cite{chang2017matterport3d} is a widely used environment in VLN that features 90 scenes and 10,800 panoramic RGB-D images, thus providing a close-to-real indoor navigation setting. The VLN agent interacts with MP3D through a connectivity graph that queries panoramic images.
The Room-to-Room (R2R) dataset, which is composed of 7,189 shortest-path trajectories in MP3D, is further augmented with 90 3D-mesh reconstructions in the Habitat simulator \cite{savva2019habitat}. This augmentation results in a continuous environment for VLN-CE. The spatially continuous R2R-CE dataset includes several validation data samples such as $\emph{val-seen}$ and $\emph{val-unseen}$ splits.
In particular, the $\emph{val-seen}$ split denotes the episodes with new paths and instructions from the same scenes observed during training. In addition, $\emph{val-unseen}$ split refers to the episodes with new paths, instructions, and scenes that are not observed during training.

\textbf{Training Setup.}
In the simulation environment of VLN-CE, the agent is defined as a mobile robot with a height of 1.5m with simulated panoramic RGB-D cameras. In addition, the position of the robot is considered available. The action space of the agent is set as \{forward 0.25m, turn-left 15°, turn-right 15°, stop\}. 
To capture panoramic RGB-D information in the current VLN-CE environment, 12 RGB cameras and 12 depth sensors are equipped on the agent, positioned at the height of 1.25m with a directional spacing of $30^{\circ}$ between each adjacent pair, which is the same setting typically used in VLN-CE studies \cite{an2023etpnav}.
The scan obtained from the constructed 2D LiDAR is represented as a radial occupancy map with a range of 3m, comprising 12 distinct range bins of 0.25m each and 120 heading bins of 3° \cite{krantz2022sim}. To enhance the navigation performance with collision avoidance, we fine-tuned the ETPNav model \cite{an2023etpnav} together with Safe-VLN. The iteration number of the fine-tuning is 20,000. With two NVIDIA RTX 3090 Ti GPUs, the total training time is approximately 50 hours. On the other hand, regarding the deployment of the trained Safe-VLN model, we have conducted evaluation tests on a single 3090 Ti GPU. The cycle-time during navigation 
is approximately 0.4s, which describes the average time it takes from the reception of the sensor feedback to the completion of the predicted action.

\begin{figure}
\centering 
\includegraphics[width=1.0\linewidth]{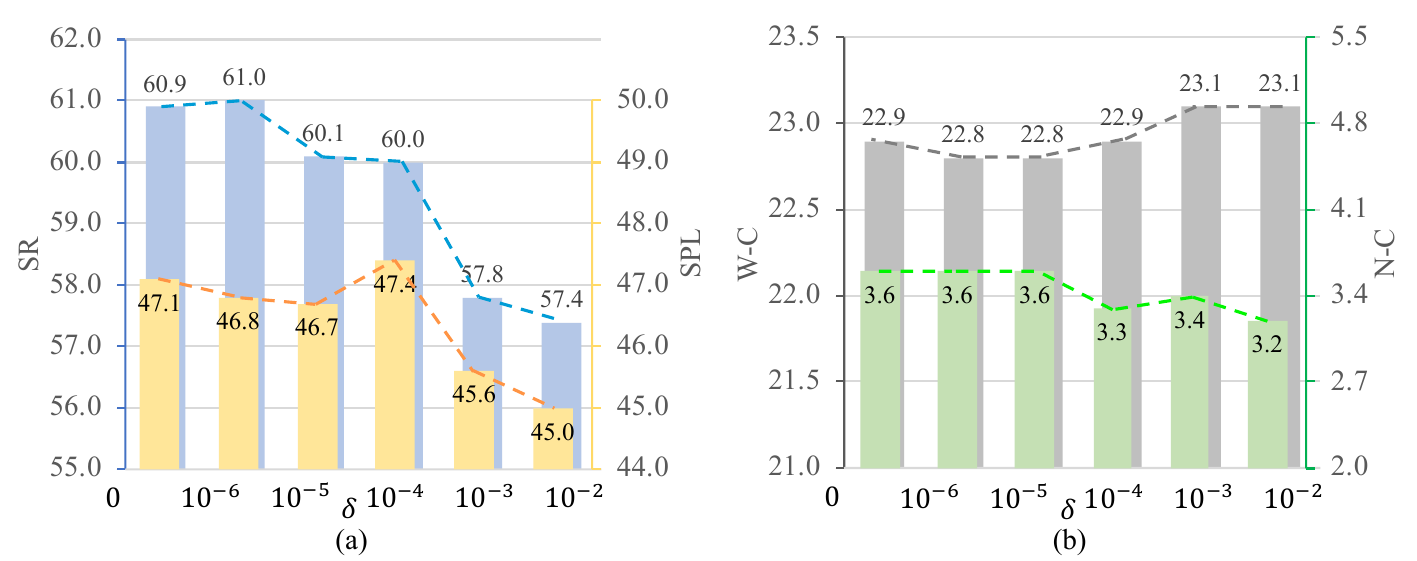}
\vspace{-0.7cm}
\caption{The experimental results of SR, SPL, N-C and W-C with varying parameter $\delta$.}
\label{delta}
\vspace{-0.75cm}
\end{figure}

\textbf{Evaluation Metrics.}
To investigate the performance of the proposed model, several evaluation metrics are adopted to compare Safe-VLN with other baselines. On one hand, following previous studies \cite{krantz2022sim,an2023etpnav}, five metrics are employed to evaluate the trained model with new paths in both seen and unseen environments. The trained model is further tested on an unseen \emph{test} split.
In particular, these metrics are (i) trajectory length (TL), i.e., the average path length; (ii) navigation error (NE), i.e., the average geometric distance between the final location and the target; (iii) oracle success rate (OSR), i.e., the rate of occurrence of paths when there exists a visited position point within 3m of the target; (iv) success rate (SR), i.e., the rate of occurrence of paths when the agent stops within 3m of the target; and (v) success rate optimized by path length (SPL).
On the other hand, to better analyze collision scenarios, three additional metrics are defined including (i) navigation collisions (N-C), i.e., the average ratio of navigation collisions in the time domain over the entire navigation process, which is calculated for the $i$-th path as $\frac{c^n_i}{T_i}$, where $c^n_i$ and $T_i$ represent the number of navigation collision steps and the total time steps, respectively; (ii) waypoint collisions (W-C), i.e., the ratio of candidate waypoints in collision area generated by the waypoint predictor, which is calculated for the $i$-th path as $\frac{1}{T_i}\sum_j^{T_i}\frac{c^w_j}{C_j}$, where $c^w_j$ and $C_j$ represent the number of generated waypoints in the collision area and the total number of generated waypoints, respectively, at the $j$-th step; and (iii) SR with 
dynamic collisions (D-C SR), i.e., the success rate when some candidate waypoints are unable to navigate due to dynamic collisions.

\subsection{Performance Comparison}\label{result}
\textbf{Comparison of VLN-CE baselines.}  We examine the effectiveness of the proposed Safe-VLN framework by testing the algorithm on three state-of-the-art VLN-CE agents, i.e., {CWP-RecBERT \cite{hong2022bridging}, GridMM \cite{gridmm} and ETPNav \cite{an2023etpnav}. This examination is conducted from two perspectives. Firstly, the three VLN-CE agents \cite{hong2022bridging, gridmm, an2023etpnav} are validated by directly adding the Safe-VLN algorithm without further training.} As indicated in Table \ref{table1}, during the validation stage, Safe-VLN enhances the success rate (SR) and effectively reduces navigation collisions (N-C), as well as waypoint collisions (W-C). In consideration of dynamic collisions, the waypoint selected by the agent is designated as non-navigable with a certain probability (10\% in this study), compelling the agent to re-select an alternate waypoint. We observe that Safe-VLN enhances the robustness of agents, thereby reducing the side effects of dynamic collisions on SR and improving dynamic collision success rate.
It is worth noting that the `Tryout' mechanism is employed during the training and validation stages of ETPNav in this study. If the `Tryout' attempt fails, the agent adjusts its action according to the re-selection navigator.
Subsequently, we further trained and validated the ETPNav agent within the Safe-VLN framework. As shown in Table \ref{table3}, the proposed Safe-VLN consistently surpasses all the VLN-CE baselines. 

\begin{table}[htbp]
\vspace{-0.2cm}
  \centering
  \caption{\centering{Validation performance of different collision avoidance methods on the R2R-CE $\emph{val-unseen}$ dataset}}
  \vspace{-0.2cm}
  \setlength{\tabcolsep}{4mm}{
    \begin{tabular}{c|c|c|c|c}
    \Xhline{1.25pt}
    \multicolumn{1}{c|}{Method} & SR    & SPL  & W-C  & N-C \bigstrut[t]\\ \hline
    \multicolumn{1}{c|}{ETPNav} & 57.1  & 49.0  & 23.3  & 14.0  \bigstrut[t]\\
   ETPNav w/ JPS & 44.8  & 26.4  & 25.9  & 17.3  \bigstrut[t]\\ 
    Safe-VLN (Ours) & 57.9$\uparrow$ & 47.8  & 22.8$\downarrow$ & 4.3$\downarrow$ \bigstrut[t]\\ 
    \Xhline{1.25pt}
    
    \end{tabular}%
  \label{jps_table}%
  }
  \vspace{-0.2cm}
\end{table}%
\textbf{Comparison with the JPS-based navigator.}
To demonstrate the effectiveness of the proposed Safe-VLN method in collision avoidance, this letter selects the widely-used JPS algorithm \cite{JPS} and conducts a comparison study. JPS is well-known as a benchmark navigator for collision issues due to
its rapid speed in navigable-path searching \cite{JPS2}. To implement JPS, a grid map of size 50×50 in an egocentric view is generated at each step by projecting
and discretizing the observed depth maps. 
The
same model parameters as in ETPNav are used in the comparison study to evaluate
three different methods, i.e., ETPNav, ETPNav in conjunction with the JPS navigator (denoted as `ETPNav w/ JPS'), and Safe-VLN on the val-unseen spit
of the R2R-CE dataset.

From the experimental result of Table \ref{jps_table}, we observe that in all the four evaluation metrics, Safe-VLN outperforms 
`ETPNav w/ JPS'. We conjecture that the performance difference comes from the projection errors when generating the grid maps used in the JPS algrithm.
For instance, in the comparison study, a portion of a planned path occasionally resides in the obstacle areas due to the projection errors near the boundary of the obstacles; and in some other cases, the goal positions are erroneously projected into the non-navigable area, which inevitably results in the malfunctioning of the JPS-based navigator. 
While this projection error will be possibly reduced if a finer grid is adopted, the resulting increased computation cost will be unaffordable. As a reference, in the comparison study presented here,
when evaluated on the val-unseen spit of the R2R-CE dataset, the time spent using the JPS-based navigator is eight times longer than that using Safe-VLN. Therefore, considering the overall performance in computational efficiency and collision avoidance success rate, Safe-VLN is superior to the widely-used JPS algorithm in the VLN-CE tasks.

\begin{table}[htbp]
  \centering
\vspace{-0.2cm}  \caption{Validation performance of Safe-VLN using LiDARs with different types and installation heights on the val-unseen split of the R2R-CE dataset}
  \vspace{-0.2cm}
    \centering
    \begin{tabular}{c|c|c|c|c|c|c}
    \Xhline{1.25pt}
    LiDAR Type & Sensor height & SR    & SPL   & W-C   & N-C & $p_o$ \bigstrut\\
    \hline
    \multirow{2}[2]{*}{2D} &1.0m & 58.2  & 48.2  & 22.5  & 4.2 & 0.384\bigstrut[t]  \\
     &1.5m & 57.9  & 47.8  & 22.8  & 4.3 & 0.377 \\
     \hline
    \multirow{2}[2]{*}{3D} &1.0m & 57.5  & 47.6  & 22.6  & 4.3 & 0.491 \bigstrut[t] \\
    &1.5m & 58.5  & 48.4  & 22.7  & 4.1 & 0.431 \\
    \hline
    \multirow{2}[2]{*}{2D+3D} & 1.0m + 1.5m & 57.7  & 47.8  & 22.7  & 4.3 & 0.439 \bigstrut[t]\\
    & 1.5m + 1.0m & 57.5  & 47.6  & 22.6  & 4.3 & 0.494\\
    \Xhline{1.25pt}
    \end{tabular}%
  \label{LiDAR}
  \vspace{-0.4cm}
\end{table}%

\begin{table*}[tb]
\vspace{-0.2cm}
  \centering
  \caption{Ablation experiments of Safe-VLN on the R2R-CE $\emph{val-unseen}$ dataset}
  \vspace{-0.2cm}
    \begin{tabular}{c|c|c|c|c|c|c|c|c|c|c|c|c|c}
    \Xhline{1.25pt}
    \multicolumn{2}{c|}{\multirow{2}[3]{*}{Training Method}} & \multicolumn{2}{c|}{\multirow{2}[3]{*}{Validation Method}} & \multicolumn{10}{c}{Validation} \bigstrut\\
\cline{5-14}    \multicolumn{2}{c|}{} & \multicolumn{2}{c|}{} & \multicolumn{5}{c|}{Val-seen}                 & \multicolumn{5}{c}{Val-unseen} \bigstrut\\
    \hline
    M-Use & R-Use & M-Use & R-Use &  SR  & SPL   & W-C  & N-C & D-C SR  &  SR   & SPL    & W-C & N-C & D-C SR \bigstrut\\
    \hline
    \multirow{4}[2]{*}{\ding{51}} & \multirow{4}[2]{*}{\ding{55}} & \ding{55}      & \ding{55}        & 68.1      & 60.6      &  22.0     & 12.1      & 65.9       &  59.2     & 48.7      &  23.4     & 10.9      & 56.7  \bigstrut\\
          &       & \ding{51}      & \ding{55}        &  68.5     & 60.8      &  22.3     & 12.2      & 67.5      & 59.3      & 48.7      &  23.3     & 10.8      & 56.1  \\
          &       & \ding{55}      & \ding{51}       &  69.8     & 60.1      & 21.0      & 3.4      & 66.5     & 59.4      & 47.2      & 23.0      & 4.3      & 56.0  \\
          &       & \ding{51}      & \ding{51}      &  69.5     &  59.4     & 21.3      & 3.6      & 66.1      & 59.9      &  48.0     &  22.7     & 4.1      & 55.5  \bigstrut\\
    \hline
    \multirow{4}[2]{*}{\ding{55}} & \multirow{4}[2]{*}{\ding{51}} & \ding{55}      & \ding{55}      & 70.1      & 60.5      & 21.7      & 7.0      & 66.6      & 56.7      & 46.7      & 23.4      & 10.8      & 54.6  \bigstrut\\
          &       & \ding{51}      & \ding{55}      & 68.9      & 60.6      &  21.7     & 6.0      & 66.6      & 58.1      & 48.0      & 23.2      & 8.5      & 55.7 \\
          &       & \ding{55}      & \ding{51}    & 68.8   & 58.9     &  20.8     & 3.1      & 67.1     & 58.8      & 46.5       & 23.0      & 3.8      & 56.4 \\
          &       & \ding{51}      & \ding{51}     & 69.0      & 59.4      & 20.9      & 2.2      &  66.2     & 57.9      & 47.2      & 22.8      & 3.4      & 54.9 \bigstrut\\
    \hline
    \multirow{4}[2]{*}{\ding{51}} & \multirow{4}[2]{*}{\ding{51}} & \ding{55}      & \ding{55}         & 68.9      & 59.8      &  21.4     & 11.3      & 66.8      & 60.0      & 47.7      & 23.2      & 9.7      & 57.7  \bigstrut\\
          &       & \ding{51}      & \ding{55}      & 69.7      & 60.2      &   21.6    &  11.2     &  67.1    & 60.0      & 47.6      & 23.1      &  9.5     & 57.6 \\
          &       & \ding{55}      & \ding{51}     & 69.3      & 58.4      &  20.8     & 3.2      &  67.6     & 60.9      & 47.1     & 22.9      & 3.6      & 57.2 \\
          &       & \ding{51}      &  \ding{51}      & 70.7      & 59.9      & 21.2      & 2.6      & 67.1      & 60.0      & 47.4     & 22.9      & 3.6      &  56.2\bigstrut\\
    \Xhline{1.25pt}
    \end{tabular}%
  \label{table2}%
\end{table*}%

\begin{figure*}[t]
\centering 
\includegraphics[width=0.9\linewidth]{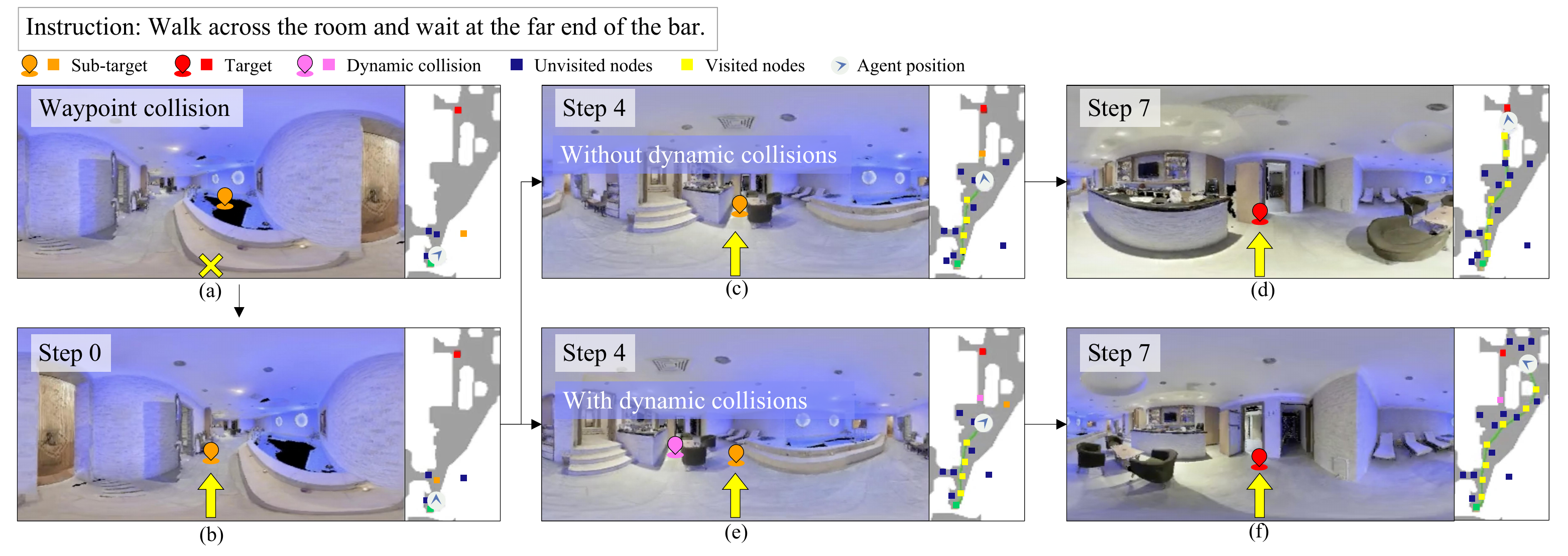}
\vspace{-0.4cm}
\caption{Examples of re-selection navigator. When the chosen waypoint is non-navigable, the agent flexibly selects another proper waypoint.}
\label{example}
\vspace{-0.6cm}
\end{figure*}
\subsection{Ablation Study}\label{ablation}


\textbf{Effectiveness of the Occupancy Mask.}
To investigate the effectiveness of the occupancy mask, we compare the performance with and without M-Use during the training stage. For the sake of brevity in the following descriptions, the use of the occupancy mask is denoted as `M-Use' and the application of the re-selection navigator is denoted as `R-Use'.
The experimental results presented in Table \ref{table2} indicate that the use of the occupancy mask consistently enhances the navigation performance across both training and validation stages. For example, the model trained with `M-Use' and validated with the full Safe-VLN (as shown in Row 4 of Table \ref{table2}) improves the success rate (SR) by 2.8\%.
The occupancy mask contributes to this improvement by providing the navigation planner with a more diverse and navigable set of candidate waypoints via the new heatmap, $H^*_t$. Specifically, as depicted in the first and third images in Fig. \ref{heat-map}, the occupancy mask, derived from the simulated 2D-LiDAR, filters out waypoints located in collision areas by reducing their sampling probability. As a result, the sampling probability in collision-free areas increases after normalization.
However, it is worth noting that due to inaccuracies in the simulated 2D-LiDAR measurement, caused by environmental reconstruction errors, occupancy masks occasionally cover free areas and filter out navigable waypoints. To balance the benefits and errors of occupancy masks, we have introduced a weight parameter, $\delta$, in Eq. (\ref{eq5}). Ablation experiments conducted on the $\emph{val-unseen}$ split under Safe-VLN with varying $\delta$ values are depicted in Fig. \ref{delta}. When $\delta$ varies within a range of less than $10^{-4}$, waypoint collisions (W-C) and navigation collisions (N-C) are relatively small, while SR and success rate optimized by path length (SPL) are high. In contrast, when $\delta$ exceeds $10^{-3}$, W-C significantly increases while SR and SPL significantly decrease. Taking into account the comprehensive navigation performance, we have applied $\delta = 10^{-4}$ in this research.


\textbf{Performance based on different LiDARs.} 
We further analyze the impact of different
types of LiDARs with different installation heights on the occupancy mask and navigation performance.
To simulate the navigation setup in a real-world environment, 2D and 3D LiDARs are constructed by combining outputs of multiple depth sensors.  The scanning range of the constructed LiDAR is 3m with a horizontal angular resolution of $0.25^{\circ}$. The 3D LiDAR possesses a vertical field of view of $22.5^{\circ}$. The minimum depth value at each horizontal scanning angle among all the vertical scanning angles is defined as the distance from the
robot to the nearest obstacle in that particular direction.  Considering the obstacle information recognized by the LiDAR varies when the LiDAR is set at different heights, we have selected two different heights of 1.0m and 1.5m as comparison examples. Furthermore, a sensor fusion strategy is also adopted and compared by combining the readings of 2D and 3D LiDARs whose sensor-fusion-based
occupancy mask is calculated by $M^{\text{2D+3D}}_t=\text{norm}(M^{\text{2D}}_t + M^{\text{3D}}_t)$, where norm$(\cdot)$ is the same normalization operation as described in Eq. (3). $M^{\text{2D}}_t$ and $M^{\text{3D}}_t$ are the occupancy masks obtained by the 2D and 3D LiDARs at the $t$-th step, respectively. 
To quantify the differences between the occupancy masks constructed from the 2D and 3D LiDARs, we calculate the average occupied proportion $p_o$ of all episodes in the val-unssen split dataset, and for the $i$-th episode, the occupied proportion is calculated by $p^{i}_o=\frac{1}{T_i}\sum_{t=1}^{T_i}(-\text{sum}(M_t)/\text{dim}(M_t))$, where $\text{dim}(\cdot)$ represents the dimension of a matrix, $T_i$ represents the total step number of the $i$-th episode, and $M_t$ represents the occupancy mask matrix at the $t$-th step.

The ablation testing results are shown in Table \ref{LiDAR}. 
Compared to the 2D LiDAR, the adoption of a 3D LiDAR enhances collision avoidance performance with more detected collision information at the height of 1.5m. 
However, the use of a 3D LiDAR at the height of 1.0m or the fused 2D and 3D LiDARs, while providing more obstacle information, actually degrade the navigation performance.
We conjecture this counter-intuitive phenomenon comes from the measurement errors of the reconstructed 3D LiDAR. 
Specifically, the low-lying furniture in the environment introduces additional errors during reconstruction due to factors such as feature occlusion and insufficient lighting, leading to a poorer
performance of LiDARs located at lower positions. In addition, we also observe that the 3D LiDAR with a wider vertical field of view, accumulates more measurement errors from the increased vertical scanning angles, especially compared to 2D LiDARs, resulting in more severe performance losses.

\textbf{Effectiveness of the Re-selection Navigator.}
We also investigate the influence of the re-selection navigator within the Safe-VLN framework.
The results demonstrate that when both `M-Use' and `R-Use' are utilized during the evaluation phase, navigation collisions (N-C) decrease to their lowest point, while the success rate (SR) experiences the most significant improvement. The re-selection navigator assists the agent in a timely manner to escape from stuck areas by enabling the re-selection of an alternate suitable waypoint.
Additionally, when `R-Use' is applied during the training phase, the gap between dynamic collision success rate (D-C SR) and SR is minimal, indicating that the agent possesses enhanced navigation flexibility for dynamic collision avoidance.
It is noteworthy that the hyper-parameters $\lambda_1$ and $\lambda_2$ in Eq. (\ref{equ6}) impact the performance of the model trained with `R-Use'. When $\lambda_1$ exceeds $\lambda_2$, the agent primarily learns about optimal actions. In contrast, sub-optimal actions are given major consideration. Taking into account the frequency of collisions, we adopt $\lambda_1=0.8$ and $\lambda_2=0.2$ in this research, with the expectation that the navigation planner will focus more on optimal actions, thereby leading to inadequate learning of sub-optimal actions. 
To provide an intuitive understanding of the effectiveness of the re-selection navigator in scenarios involving waypoint collision and dynamic collision, we present an example in Fig. \ref{example}. As shown in Fig. \ref{example}(a), the agent initially selects the predicted waypoint located in a non-navigable pool as the sub-target. The agent then adjusts its sub-target waypoint by re-selecting a new navigable candidate waypoint (as shown in Fig. \ref{example}(b)) and successfully reaches the target (as shown in Figs. \ref{example}(c)-(d)). Moreover, as depicted in Fig. \ref{example}(e), to avoid dynamic collisions, the agent chooses an alternate path to reach the target (as shown in Figs. \ref{example}(e)-(f)). This example demonstrates that with the re-selection navigator, VLN-CE agents successfully and efficiently avoid collisions in continuous environments.


\section{Conclusion}\label{conclusion}

This letter presented the comprehensive analysis of collision avoidance in the field of VLN-CE, laying the foundation for the practical deployment of VLN in real-world applications. We examined collision scenarios in VLN-CE and their corresponding influences on the navigation performance. We proposed a collision-avoidance framework named Safe-VLN. The proposed method employed occupancy masks to guide the waypoint predictor towards generating candidate waypoints in collision-free areas. Additionally, Safe-VLN incorporated a re-selection navigator to encourage agents to adapt their actions flexibly, thus preventing them from becoming trapped in collision areas. 
We plan to deploy the proposed Safe-VLN to a real-world robot
and conduct collision avoidance evaluation and analysis. To achieve this,
we will investigate the sim-to-real transfer of Safe-VLN and explore various methods to enhance the real-world environmental perception such as augmenting the training data with real-world observations. 

\addtolength{\textheight}{-12cm}   





\bibliographystyle{./IEEEtran} 
\bibliography{./reference}
\end{document}